\pdfoutput=1

\documentclass[11pt]{article}

\usepackage[final]{acl}

\usepackage{times}
\usepackage{latexsym}

\usepackage[T1]{fontenc}

\usepackage[utf8]{inputenc}

\usepackage{microtype}

\usepackage{inconsolata}

\usepackage{graphicx}

\usepackage{amsmath}
\usepackage{multirow}
\usepackage{amssymb}

\usepackage{ makecell} 
\usepackage{booktabs} 
\title{The Role of Background Information in Reducing Object Hallucination in Vision-Language Models: Insights from Cutoff API Prompting}

\author{Masayo Tomita \and Katsuhiko Hayashi \and Tomoyuki Kaneko \\
The University of Tokyo \\
\texttt{mtomita143@gmail.com}\ \ \ \ \ \texttt{katsuhiko-hayashi@g.ecc.u-tokyo.ac.jp} \\
\texttt{kaneko@graco.c.u-tokyo.ac.jp}}

\begin{document}
\maketitle
\begin{abstract}
Vision-Language Models (VLMs) occasionally generate outputs that contradict input images, constraining their reliability in real-world applications. While visual prompting is reported to suppress hallucinations by augmenting prompts with relevant area inside an image, the effectiveness in terms of the area remains uncertain. This study analyzes success and failure cases of Attention-driven visual prompting in object hallucination, revealing that preserving background context is crucial for mitigating object hallucination.

\end{abstract}

\section{Introduction}
Vision-Language Models (VLMs), represented by CLIP~\cite{clip} and LLaVA~\cite{llava}, have advanced rapidly and garnered attention. A Vision-Language Model is equipped with an image encoder and a text encoder, allowing it to process both image and text information simultaneously. With these encoders, multimodal tasks such as visual question answering~\cite{vqa}, image captioning~\cite{caption}, and image retrieval~\cite{retrieval} have been realized, and applications have expanded into fields like autonomous driving~\cite{autonomous} and medical diagnosis~\cite{medical}. However, as these applications expand, the issue of hallucination has become increasingly apparent.

Hallucination refers to inconsistencies between the input image and the output text~\cite{hallucination}. Among them, Hallucinations of an object's presence, attributes, and relationships are defined as object, attribution, and relation hallucinations, respectively. POPE~\cite{pope} has been proposed and widely used as a method to evaluate the object hallucination. POPE works by asking questions about objects that are either present or absent in an image and verifying whether the responses are correct~\cite{pope}.

Prompting in VLMs has been proposed as methods to reduce hallucination~\cite{prompting}. In particular, visual prompting, which emphasizes target objects by adding marks has been shown to suppress hallucination~\cite{visualprompting}. Among these techniques, a method of visual prompting known as Attention Prompting on Image (API) has been introduced; it is automatic way of visual prompting, which demonstrated improved accuracy in VQA and object hallucination~\cite{api}. 

\begin{figure}[t]
    \centering
    \includegraphics[width=0.45\textwidth]{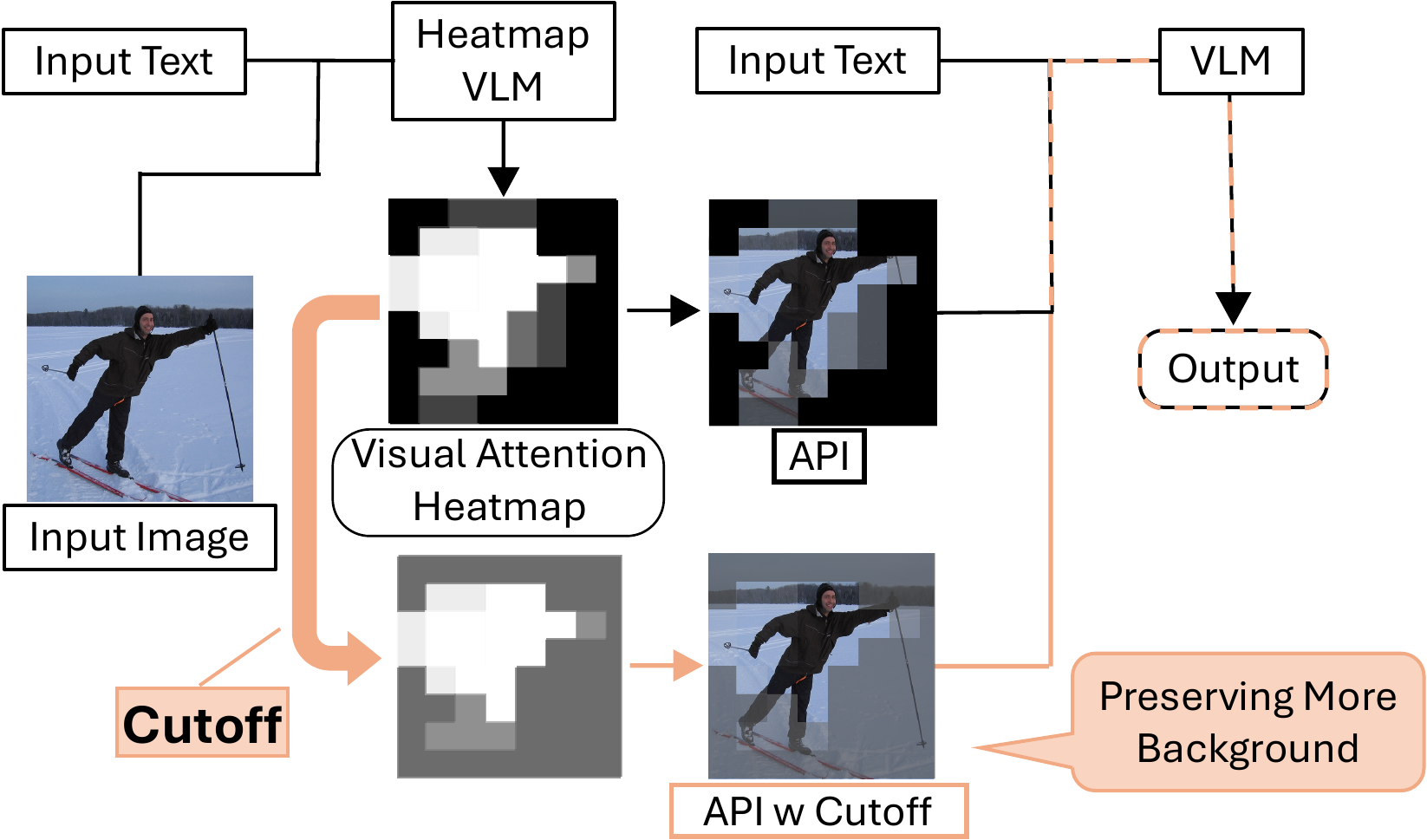}
    \caption{API Prompting Process: Without Cutoff~\cite{api} vs. With Cutoff (Proposed).}
    \label{api}
\end{figure}

However, in the work reported by~\citet{api}, experiments mostly focused on VQA, leaving a deeper analysis of its effect on object hallucination unexamined. This study analyzes success and failure cases of API prompting to explore effective visual prompting for reducing object hallucination. Specifically, this work examines the relationship between attention-target object alignment and API prompting accuracy and evaluates whether removing the background reduces object hallucination as a validation of cases where attention was focused solely on the target object.

\section{Related Work}
Visual prompting techniques is reported to enhance VQA accuracy and reduce hallucinations by emphasizing key image regions. For example, the Engineering Visual Prompting method addes red circles and arrows by hand~\citep{redcircle, mark}, while a semi-automated approach overlays segmentation data from models such as SEEM~\citep{seem}. To experiment on over 20,000 objects, API prompting, which is fully automated approach, was employed.


\section{Method}
\begin{figure}[t!]
    \centering
    \includegraphics[width=0.45\textwidth]{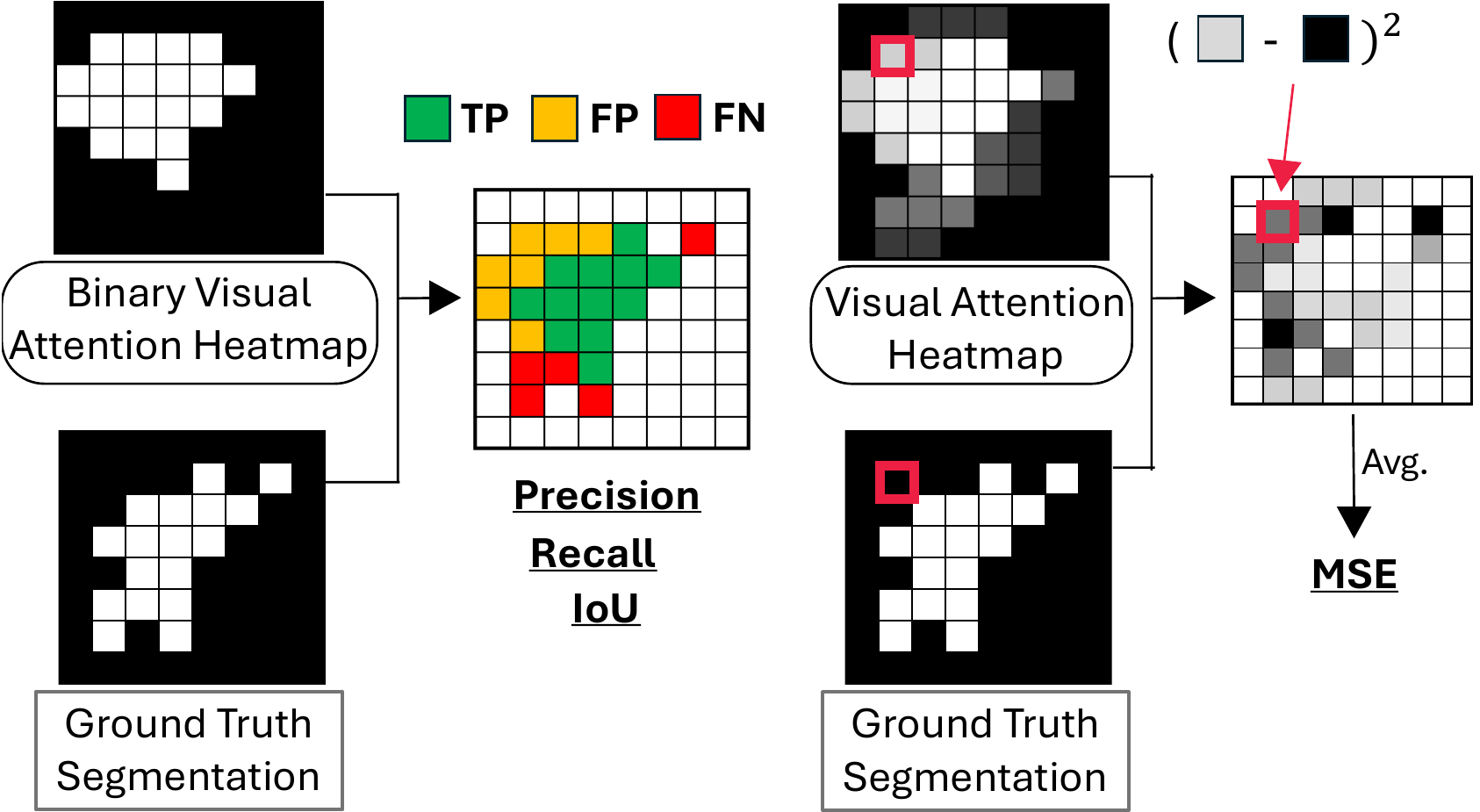}
    \caption{Process of Visual Attention Evaluation.}
    \label{img1}
\end{figure}

\subsection{API Prompting}
API Prompting~(sketched in Figure~\ref{api}) is a Visual Prompting method that highlights important parts in an image using a Visual Attention Heatmap derived from a Vision-Language Model~\citep{api}. The Attribution Map, representing the contribution of image tokens to model outputs, is extracted from a VLM (referred to as Heatmap VLM or H-VLM), convolved, resized to match the image size, and then overlaid on the original image.

Following the study by \citet{api}, Vision-Transformer-based CLIP and LLaVA are employed as Heatmap VLMs. The methods for extracting Visual Attention Attribution Maps from each model are described below.

\paragraph{CLIP Attribution Map}
CLIP computes similarity between text and image representations, and the Attribution Map \( \Psi \) is obtained by decomposing the similarity function \( \text{sim}(\hat{I}, \hat{T}) \). 

Since later MSA layers greatly impact image representation~\citep{clipdec}, the similarity function is approximated as:
\begin{equation}
   \begin{aligned}
\text{sim}(\hat{I}, \hat{T}) \approx \text{sim}\left(\sum_
{\ell=L'}^{L} \mathcal{L}(\text{MSA}^{\ell}([Z^{\ell-1}]))_{\text{cls}}, \hat{T}\right).
\end{aligned}
\end{equation}
To filter out irrelevant regions, a complementary Attribution Map \( \Psi^{\text{comp}} \) is introduced:
\begin{equation}
   \begin{aligned}
   \Psi_{i,j}^{\text{comp}} &\triangleq 1 - \text{sim}(\mathcal{L}(Z_{t}^{L}), \hat{T}), \\
&\quad \text{where} \quad t = 1 + j + P \cdot (i - 1).
\end{aligned}
\end{equation}
Combining both maps, the final CLIP Attribution Map is defined as:
\begin{equation}
   \begin{aligned}
   \Psi = \Psi^{\text{cls}} + \Psi^{\text{comp}} - \Psi^{\text{comp}} \odot \Psi^{\text{cls}}.
\end{aligned}
\end{equation}

\paragraph{LLaVA Attribution Map}
LLaVA can provide an Attribution Map \( \Psi \) using Multi-Head Self-Attention (MSA) weights between output text tokens and image tokens. The Attribution Map is computed by averaging over all output tokens and attention heads:
\begin{equation}
   \begin{aligned}
\Psi_{i,j} &\triangleq \frac{1}{MH} 
  \sum_{m=1}^{M} 
  \sum_{h=1}^{H} 
  A_{m,t}^{(\bar{L},h)}, \\
&\quad \text{where} \quad 
t = j + P \cdot (i - 1).
\end{aligned}
\end{equation}
Here, \(M\) is the number of output tokens, \(H\) is the number of attention heads, \(P\) is the number of patches per image side, and \(A^{(\bar{L},h)}\) represents cross-attention between output text and image tokens at layer \(\bar{L}\) and attention head \(h\).

\subsection{Background Role Examination}
To assess the necessity of background information for object recognition, ground truth segmentation data is used as a Heatmap during API Prompting and the accuracy of output is evaluated (hereafter referred to as API - Seg.). Binary segmentation masks, overlaid in gray are input into VLMs to evaluate their impact on output accuracy. If POPE response accuracy remains unchanged, background information is deemed unnecessary.

\subsection{Minimum Cutoff}
Minimum cutoff redefines the minimum value in segmentation or Visual Attention Heatmap based on a threshold. Since a threshold of 0.5 showed improvement in Table~\ref{table2}, values below 0.5 in the cutoff are replaced with 0.5, refining segmentation granularity.

\begin{table*}[t]
\centering
\begin{tabular}{lllrrrrr}
\toprule
\textbf{Dataset} & \textbf{Model} & \textbf{Prompting} & \textbf{Acc.} & \textbf{Prec.} & \textbf{Rec.} & \textbf{TNR} & \textbf{F1} 
\\ \cmidrule(lr){1-8}
\multirow{7}{*}{\textbf{MSCOCO}} & \multirow{7}{*}{LLaVA} & w/o prpt. & 86.23 & 84.21  & 89.19 & 83.27 & 86.63 \\ 
& & API~(CLIP) & \ensuremath{\blacktriangle}86.52 & \ensuremath{\blacktriangle}84.78 & \ensuremath{\triangledown}89.02 & \ensuremath{\blacktriangle}84.02 & \ensuremath{\blacktriangle}86.85 \\ 
& & API~(CLIP) w Cutoff & \ensuremath{\blacktriangle}88.59 & \ensuremath{\blacktriangle}85.84 & \ensuremath{\blacktriangle}92.43 & \ensuremath{\blacktriangle}84.75 & \ensuremath{\blacktriangle}89.01 \\
 & & API~(LLaVA) & \ensuremath{\triangledown}86.11 & \ensuremath{\blacktriangle}84.72 & \ensuremath{\triangledown}88.12 & \ensuremath{\blacktriangle}84.10 & \ensuremath{\triangledown}86.39 \\ 
 & & API~(LLaVA) w Cutoff & \ensuremath{\blacktriangle}87.98 & \ensuremath{\blacktriangle}85.10 & \ensuremath{\blacktriangle}92.09 & \ensuremath{\blacktriangle}83.88 & \ensuremath{\blacktriangle}88.46 \\
  & & API - Seg. & - & - & \ensuremath{\triangledown}71.78 & - &  - \\ 
  & & API - Seg. w Cutoff & - & - & \ensuremath{\blacktriangle}89.24 & - &  - \\ 
 \bottomrule
\end{tabular}
\caption{POPE results on MSCOCO datasets with API Prompting.}
\label{table1}
\end{table*}

\begin{table*}[h!]
\centering
\begin{tabular}{llrrrr}
\toprule
\multirow{2}{*}{\textbf{H-VLM}} & \multirow{2}{*}{\textbf{Output}} & \multicolumn{4}{c}{\textbf{Visual Attention Alignment}} \\ 
& & \textbf{Prec.} & \textbf{Rec.}  & \textbf{IoU}    & \textbf{MSE}   \\ 
\cmidrule(lr){1-6}
\multirow{2}{*}{CLIP}& Correct~(87\%)  &\ensuremath{\blacktriangle}13.52   & \ensuremath{\blacktriangle}83.94 & \ensuremath{\blacktriangle}11.68 & \ensuremath{\blacktriangle}28.46  \\
& Incorrect~(13\%) & 6.09   & 77.02 & 4.78 & 35.22 \\
\multirow{2}{*}{LLaVA}& Correct~(86\%)  & \ensuremath{\blacktriangle}10.62  & 64.62 & \ensuremath{\blacktriangle}8.10 & 13.60 \\ 
& Incorrect~(14\%) & 6.17   & \ensuremath{\blacktriangle}69.78 & 4.44 & \ensuremath{\blacktriangle}11.60 \\
\bottomrule
\end{tabular}
\caption{Alignment of CLIP/LLaVA Visual Attention.}
\label{table3}
\end{table*}

\subsection{Evaluation of Visual Attention Alignment}
To assess how well visual attention focuses on target objects, Precision, Recall, Intersection over Union~(IoU), and Mean Squared Error~(MSE) are computed between the object segmentation data and the Visual Attention Heatmap as depicted in Figure~\ref{img1}. 
Visual Attention Heatmaps are converted into binary arrays using thresholds set to the average value of each heatmap.


\section{Experiment}
\subsection{Setting}
A total of 3,860 images from the MSCOCO dataset~\cite{mscoco} were used to generate visual attention heatmaps from CLIP and LLaVA. These heatmaps were overlaid onto the original images as black masks, and the masked images, along with POPE questions, were input into LLaVA to evaluate the outputs and assess visual attention.

The objects in each image were identified using MSCOCO labels, with three absent objects randomly selected from the remaining 80 labels. Segmentation data for the present objects was retrieved from MSCOCO and used as ground truth.

The API Prompting method followed the published implementation\footnote{\url{https://github.com/yu-rp/apiprompting}}.
CLIP API Prompting used a pre-trained CLIP ViT-L/14@336px model, while LLaVA API Prompting employed llava-v1.5-7b. Visual attention heatmaps were extracted from layer 22 for CLIP and layer 20 for LLaVA. A convolution with a kernel size of 3 was applied, and the maps were resized using the LANCZOS method.

POPE question responses were generated using llava-hf/llava-1.5-7b-hf (vLLM version) with a temperature of \texttt{0.8} and top\_p of \texttt{0.9}. The phrase “Answer Yes, No, or Not Sure” was appended to each question, and responses were classified based on the leading token.



\subsection{Result}
Table~\ref{table1} shows the results of API Prompting with CLIP, LLaVA, and ground truth segmentation, with and without Cutoff. Both CLIP and LLaVA API Prompting without Cutoff improved metrics such as True Negative Rate~(TNR), but the improvement remained below 1\%. Providing segmentation significantly reduced Recall. In contrast, API Prompting with Cutoff showed an improvement especially in Recall of approximately 3\%.

As shown in Table~\ref{table3}, when the output of API Prompting was correct, the alignment between the target object and visual attention was better, with a difference of approximately 5\%. Notably, the precision and IoU exhibited this trend across both Heatmap VLMs, whereas the recall and MSE improvements were observed only when using CLIP as the Heatmap VLM.

As Table~\ref{table5} summarized, in cases where the VLMs answered correctly without background information (upper section of the table), the accuracy of API Prompting without Cutoff was comparable to that of the w/o prompting condition. However, in cases where the model answered incorrectly without background information (lower section of the table), the accuracy significantly deteriorated. With Cutoff, especially in cases where the model initially answered incorrectly without background information, a substantial improvement in recall was observed compared to the w/o prompting condition.

\section{Discussion}
\begin{table}[t!]
\centering
\begin{tabular}{llr}
\toprule
\textbf{API - Seg.} & \textbf{Prompting} & \textbf{Rec.} \\ \cmidrule(lr){1-3}
\multirowcell{5}{Correct\\(72\%)} & w/o prpt. & 94.74\\
& API~(CLIP)& \ensuremath{\blacktriangle}95.02 \\
& API~(CLIP) w Cutoff & \ensuremath{\blacktriangle}97.08 \\
 & API~(LLaVA) & \ensuremath{\triangledown}94.23  \\
 & API~(LLaVA) w Cutoff & \ensuremath{\blacktriangle}96.64
 \\\cmidrule(lr){1-3}
\multirowcell{5}{Incorrect\\(28\%)} & w/o prpt. & 75.05\\
& API~(CLIP) & \ensuremath{\triangledown}73.75  \\
& API~(CLIP) w Cutoff & \ensuremath{\blacktriangle}80.61  \\
 & API~(LLaVA) & \ensuremath{\triangledown}72.59  \\
 & API~(LLaVA) w Cutoff & \ensuremath{\blacktriangle}80.51 \\
 \bottomrule
\end{tabular}
\caption{POPE results categorized by correct and incorrect API - Seg. outputs.}
\label{table5}
\end{table}

The overall better performance of API Prompting with Cutoff supports the claim that masking parts of the image is not an effective visual prompting method for reducing object hallucination. This conclusion is further reinforced by the significantly worse results of API - Seg. without Cutoff compared to the w/o prompting case.

This is likely due to the fact that background information surrounding the target object contributes to its identification. Given that API Prompting has shown particularly strong results in prior studies, especially in VQA, conventional API Prompting, which completely hides parts of the image, may be effective when the area requiring attention is already clear to the VLM. However, in the case of object hallucination, the VLM appears to identify objects in conjunction with surrounding information, making the complete removal of image content undesirable. Nevertheless, considering that API Prompting with Cutoff outperforms the w/o prompting case, emphasizing the target object appears to be effective in reducing hallucination.  

As Table~\ref{table6} demonstrated, in cases where responses were successful without background information, observing API Prompting results at an IoU threshold of 10 reveals that when IoU was greater than 10, meaning the target object and attention were well aligned, API Prompting was particularly effective. This indicates that when an answer can be derived without background information, API Prompting is more effective if the target object is emphasized.  

\begin{figure}[t]
    \centering
    \includegraphics[width=0.47\textwidth]{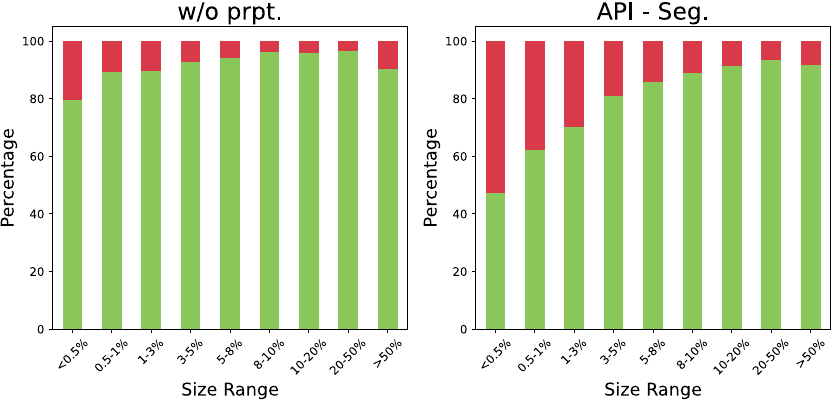}
    \caption{Image Size vs. POPE Results.}
    \label{size}
\end{figure}

As shown in Figure~\ref{size}, API - Seg. without Cutoff tends to produce incorrect outputs when the target object is small. This suggests that background information plays a more crucial role for smaller objects.

\section{Conclusion}
\begin{table}[t!]
\centering
\begin{tabular}{lllr}
\toprule
\textbf{H-VLM} & \textbf{IoU} & \textbf{Prompting} & \textbf{Rec.} \\ \cmidrule(lr){1-4}
\multirow{4}{*}{CLIP} &\multirow{2}{*}{$\geqq$ 10} & w/o prpt. & 92.60\\
&& API~(CLIP)& \ensuremath{\blacktriangle}93.16 \\
&\multirow{2}{*}{< 10} & w/o prpt. & 97.46\\
&& API~(CLIP)& \ensuremath{\triangledown}97.37 \\
 \cmidrule(lr){1-4}
\multirow{4}{*}{LLaVA} & \multirow{2}{*}{$\geqq$ 10} & w/o prpt. & 93.64\\
&& API~(LLaVA) & \ensuremath{\blacktriangle}93.65  \\
&\multirow{2}{*}{< 10} & w/o prpt. & 96.82\\
&& API~(LLaVA) & \ensuremath{\triangledown}95.02  \\
 \bottomrule
\end{tabular}
\caption{POPE results on objects present in cases where the output of API - Seg. is correct.}
\label{table6}
\end{table}

In conclusion, an analysis of success and failure cases in attention-driven visual prompting for object hallucination yielded the following insights:

\begin{enumerate}
\item Background information is essential for VLMs to accurately determine object presence.
\item Cutoff API Prompting improves performance by both highlighting relevant objects and preserving background context.
\item API Prompting is more effective when visual attention aligns well with the target objects.
\item Smaller objects rely more heavily on background information.
\end{enumerate}

Visual prompting that partially masks an image, even if limited to the background, proves ineffective in mitigating object hallucination. Conversely, emphasizing the target object without full occlusion contributes to its reduction. Future visual prompting should focus on enhancing visibility rather than obscuring image elements.

\clearpage
\section*{Limitations}
One limitation of this study is that it only evaluated LLaVA as the target Vision Language Model (VLM), which may limit the generalizability of the findings to other models. Additionally, the alignment of visual attention heatmaps for non-existing objects was not assessed, indicating that further analysis is needed in this area. 

Moreover, the experiments were conducted solely using the MSCOCO dataset, and future work should expand the evaluation to include additional datasets to ensure the robustness and broader applicability of the results. Furthermore, since datasets that contain both questions and corresponding answers alongside matching segmentation data, which can be used to evaluate object hallucination, are scarce, it may be necessary to develop such datasets.

\section*{Ethical Considerations}
This study employs the open-source MSCOCO dataset, restricting object categories to its 80 predefined classes. While this ensures consistency with previous vision-language research, it also introduces potential bias by excluding objects outside these classes. Future work should assess whether similar visual prompting techniques are effective across a broader range of objects and datasets.

Additionally, API Prompting method relies entirely on VLM outputs, constraining its performance to the accuracy and biases of the underlying model. Although this study focuses on mitigating object hallucination, API Prompting cannot guarantee fully reliable object recognition. Researchers and practitioners should be mindful of these limitations, especially in high-stakes scenarios where incorrect outputs might lead to misinterpretations.

\bibliography{ACL2025}

\clearpage
\appendix

\label{sec:appendix}
\onecolumn
\section{Detail of Minimum Cutoff}

\begin{table}[h!]
\centering
\begin{tabular}{lcr}
\toprule
\textbf{Prompting} & \textbf{Cutoff} &  \textbf{Rec.}  \\ \cmidrule(lr){1-3}
 w/o prpt & w/o Cutoff & 89.19 \\
\multirow{5}{*}{API - Seg.} & w/o & \ensuremath{\triangledown}71.78 \\
& 0.1 & \ensuremath{\triangledown}86.72 \\
& 0.2 & \ensuremath{\triangledown}88.09 \\
& 0.3 & \ensuremath{\triangledown}88.61 \\
& 0.4 & \ensuremath{\triangledown}88.77 \\
& 0.5 & \ensuremath{\blacktriangle}89.24 \\
 \bottomrule
\end{tabular}
\caption{POPE results on MSCOCO with segmentation API Prompting across cutoff values.}
\label{table2}
\end{table}

Table~\ref{table2} shows the results of API Prompting with ground truth segmentation at different cutoff values. As the clipping value increased, there was a trend toward improved results. The case where the cutoff value was 0.5 showed the best performance, outperforming the case without prompting images. Significant improvements were observed when comparing the case without cutoff and when cutoff at a value of 0.1.

\begin{figure}[h!]
    \centering
    \includegraphics[width=0.97\textwidth]{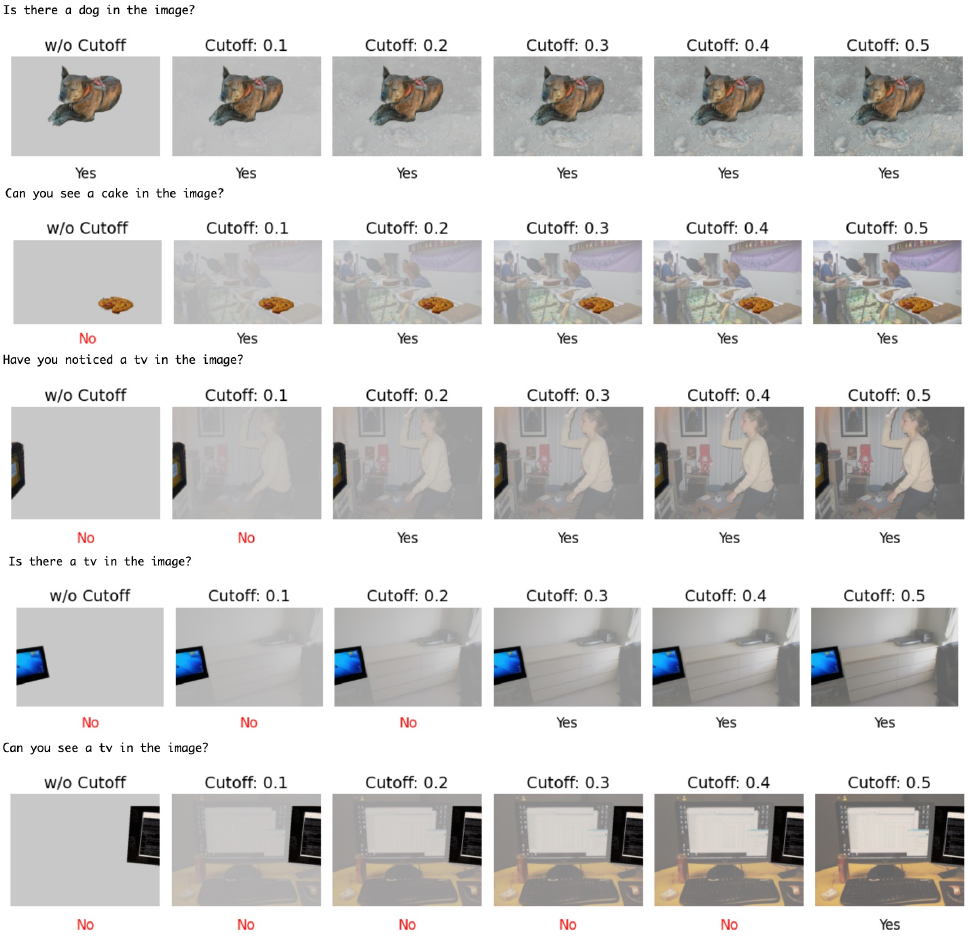}
    \caption{Examples of Cutoff Segmentation.}
\end{figure}

\newpage

\section{Successful Cases}
Examples are presented where images without prompting led to incorrect outputs, while API Prompting resulted in correct outputs.

\subsection{Objects Present}
\begin{figure}[h!]
    \centering
    \includegraphics[width=0.90\textwidth]{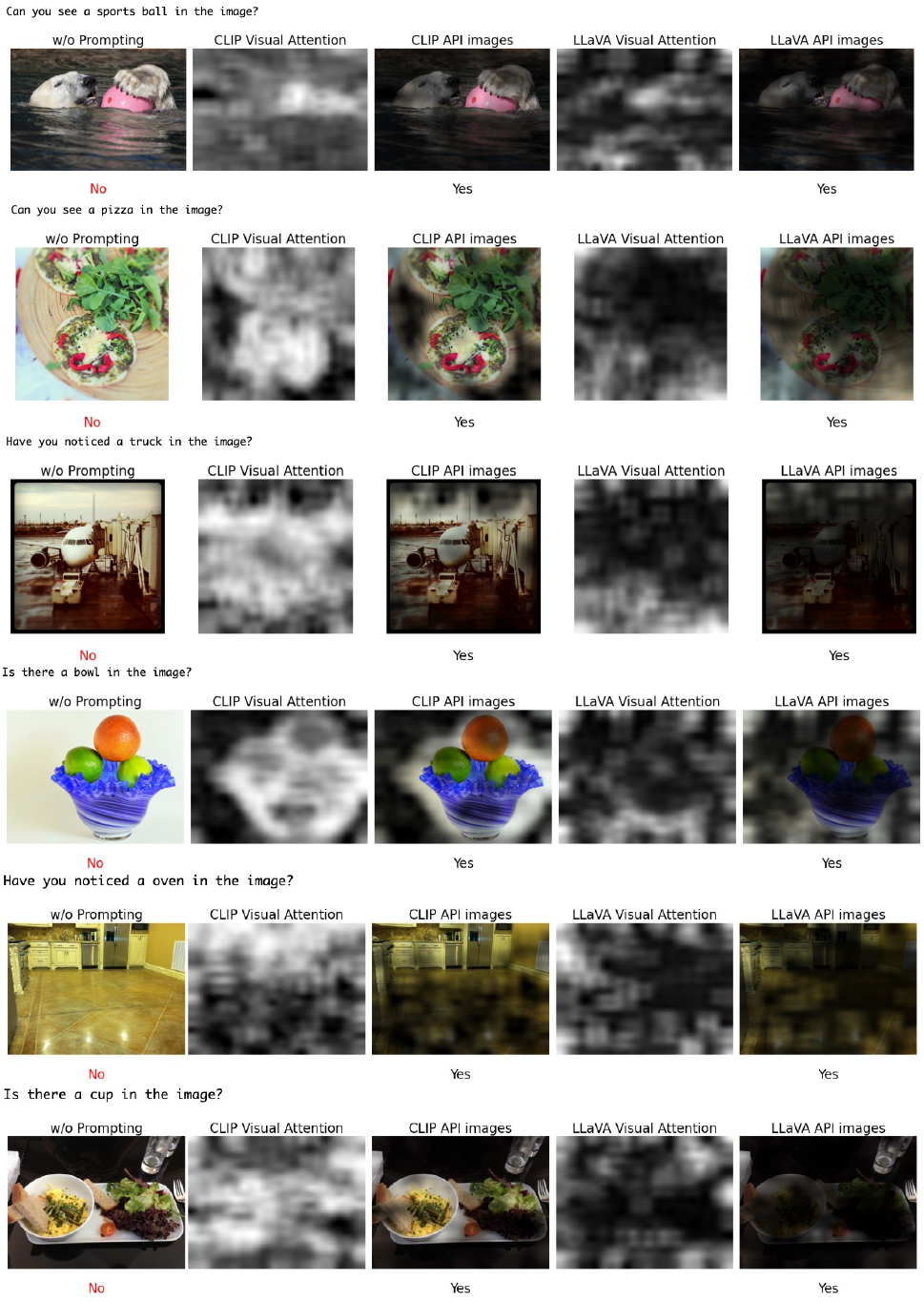}
    \caption{Examples of Successful Cases of Objects Present in Images.}
\end{figure}

\newpage
\subsection{Objects Absent}
\begin{figure}[h!]
    \centering
    \includegraphics[width=0.9\textwidth]{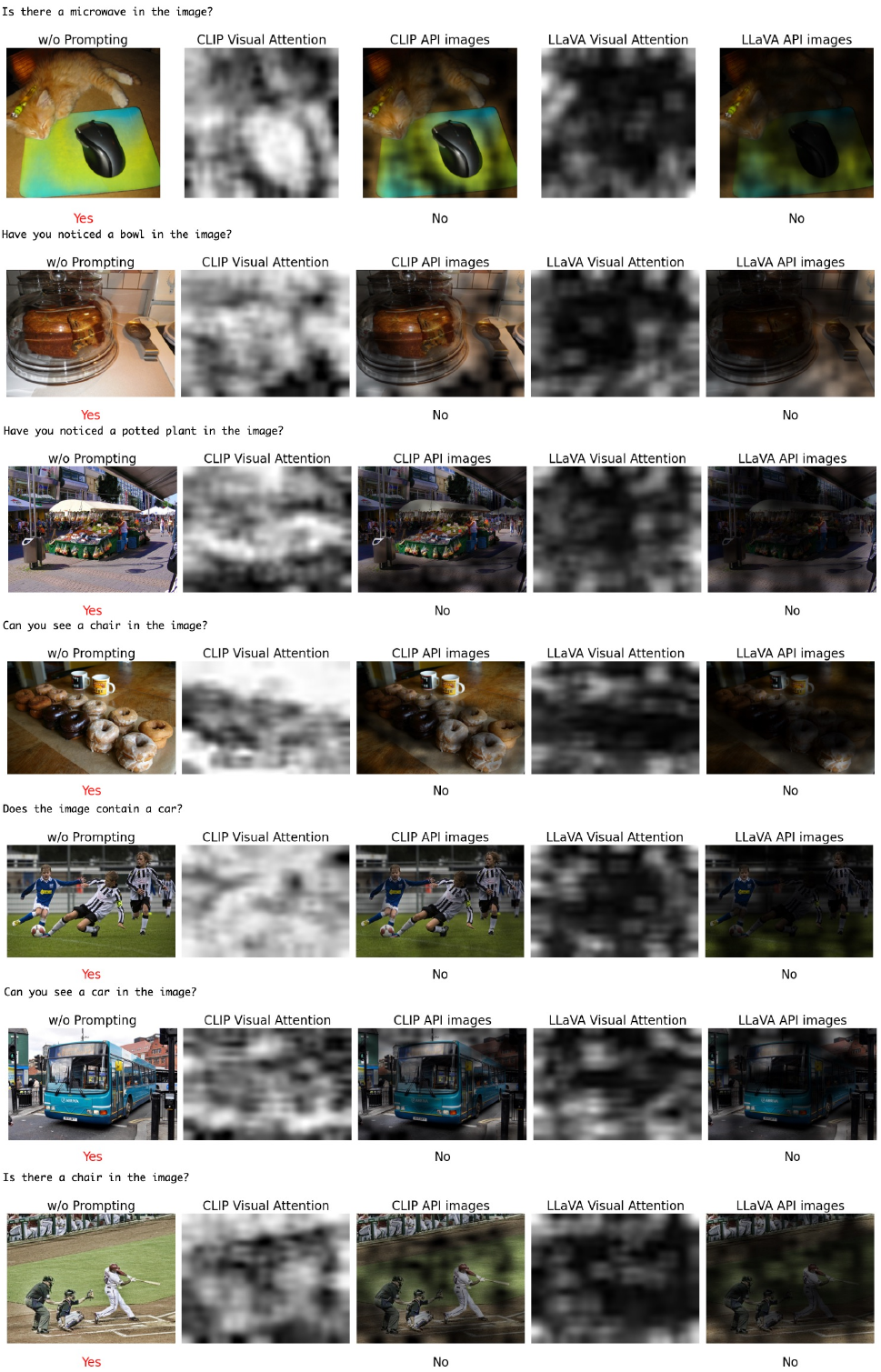}
    \caption{Examples of Successful Cases of Objects Absent in Images.}
\end{figure}

\clearpage

\section{Failed Cases}
Examples are presented where images without prompting led to correct outputs, while API Prompting resulted in incorrect outputs.

\subsection{Objects Present}
\begin{figure}[h!]
    \centering
    \includegraphics[width=1.0\textwidth]{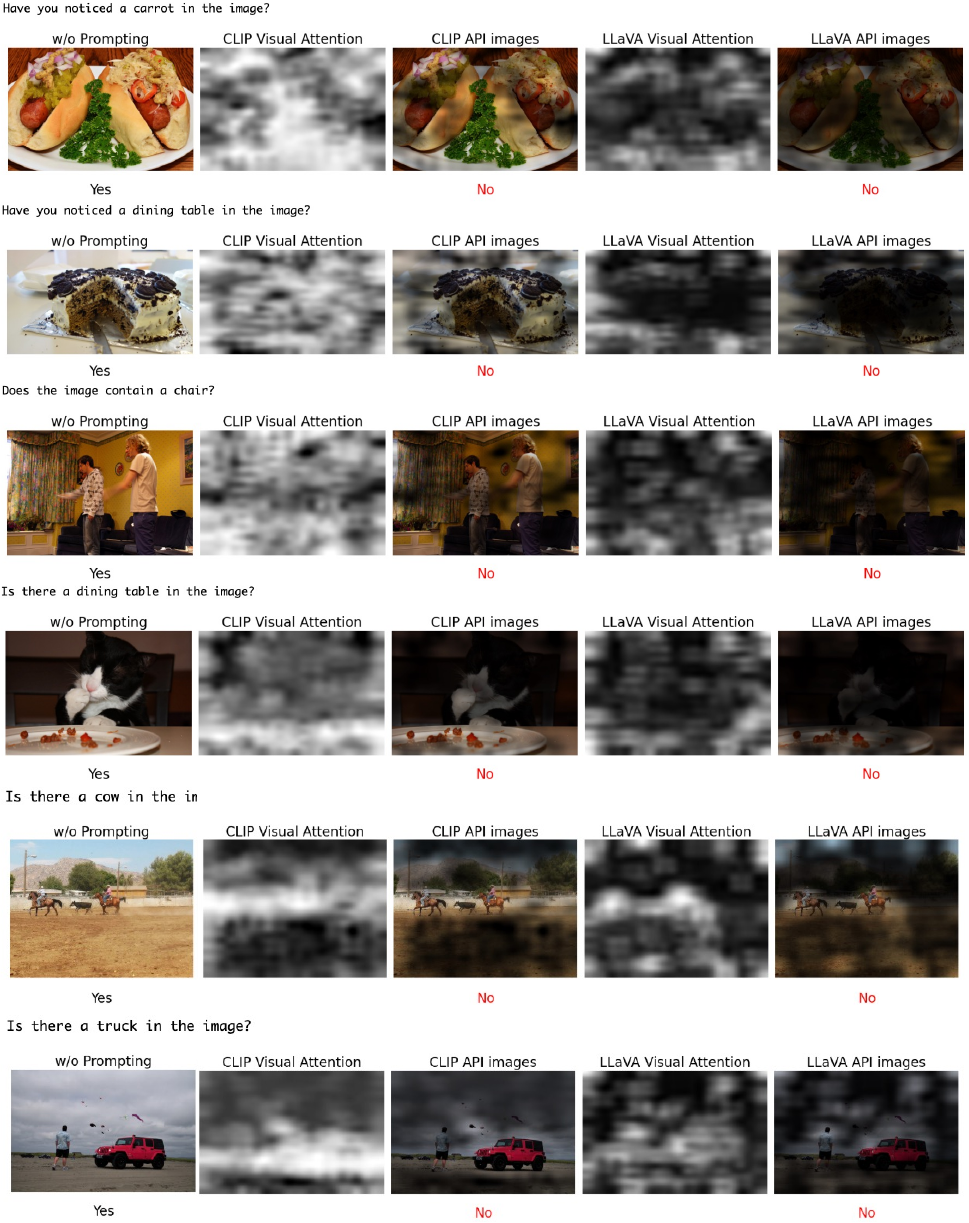}
    \caption{Examples of Failed Cases of Objects Present in Images.}
\end{figure}

\clearpage
\subsection{Objects Absent}

\begin{figure}[h!]
    \centering
    \includegraphics[width=0.95\textwidth]{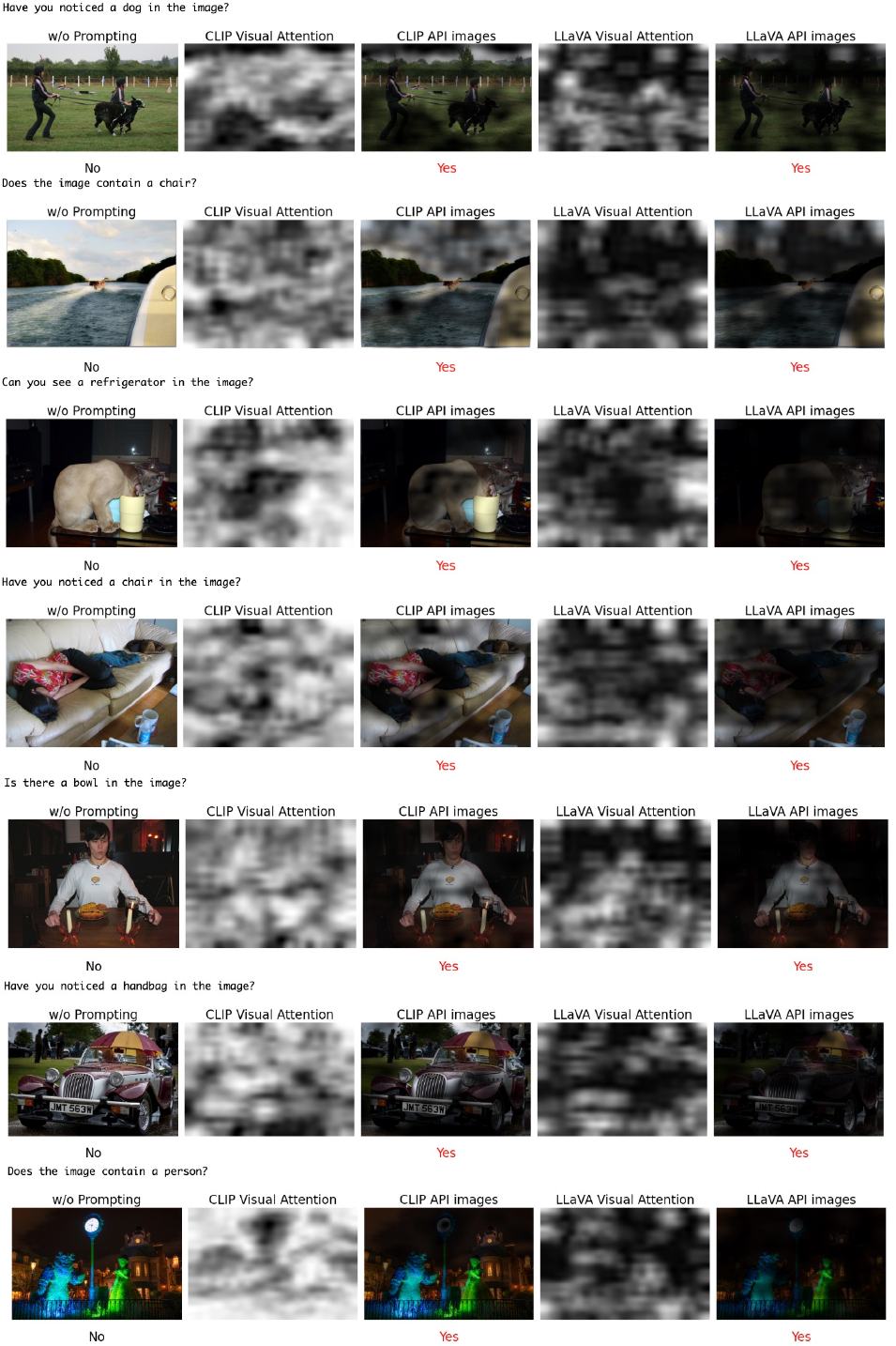}
    \caption{Examples of Failed Cases of Objects Absent in Images.}
\end{figure}
\newpage

\section{Cutoff Successful Cases}
Examples are presented where API Prompting images without cutoff led to incorrect outputs, while Cutoff API Prompting resulted in correct outputs.

\subsection{Objects Present}
\begin{figure}[h!]
    \centering
    \includegraphics[width=0.96\textwidth]{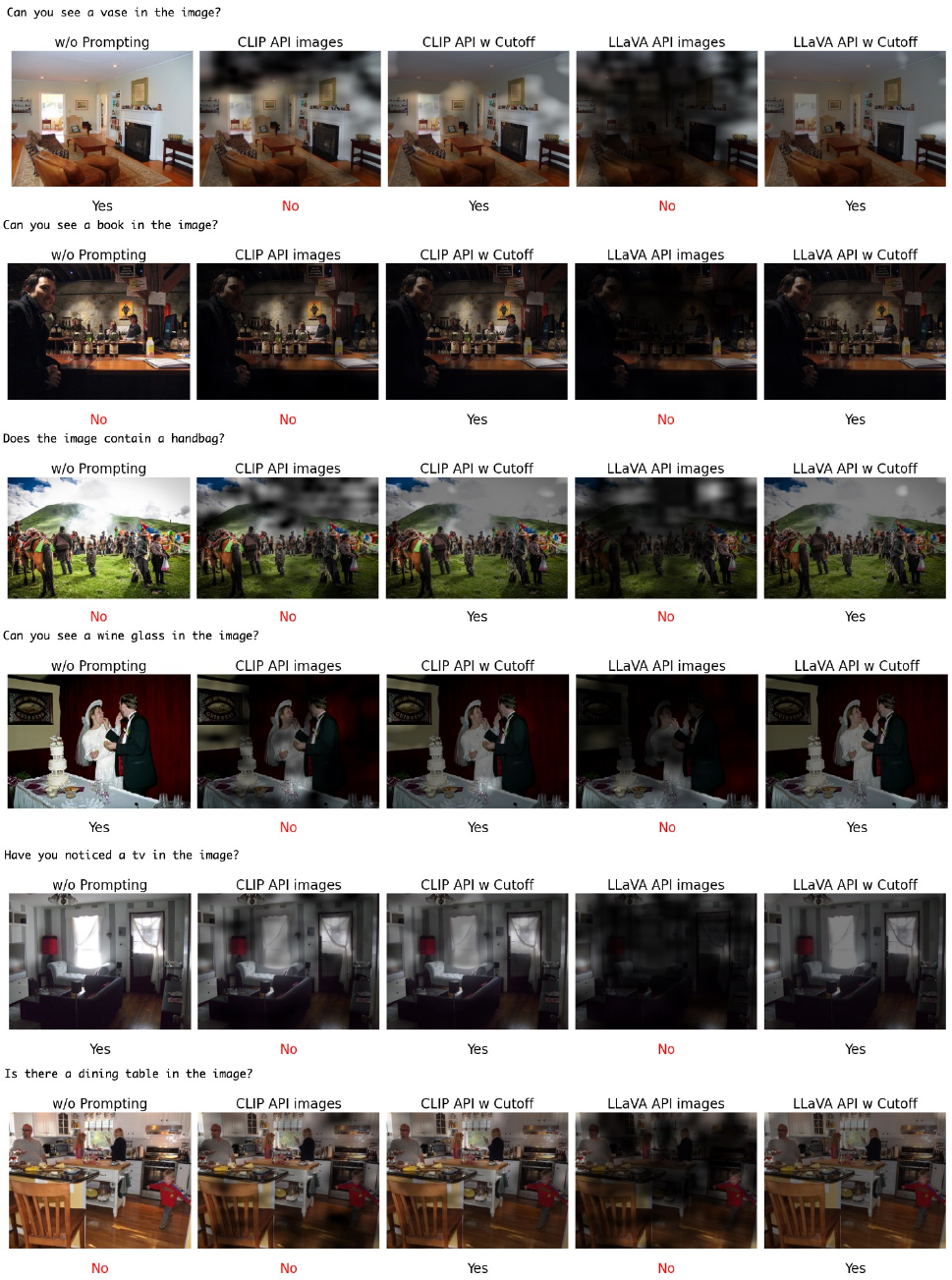}
    \caption{Examples of Successful Cases of Cutoff API Prompting of Objects Present in Images.}
\end{figure}

\clearpage
\subsection{Objects Absent}
\begin{figure}[h!]
    \centering
    \includegraphics[width=0.96\textwidth]{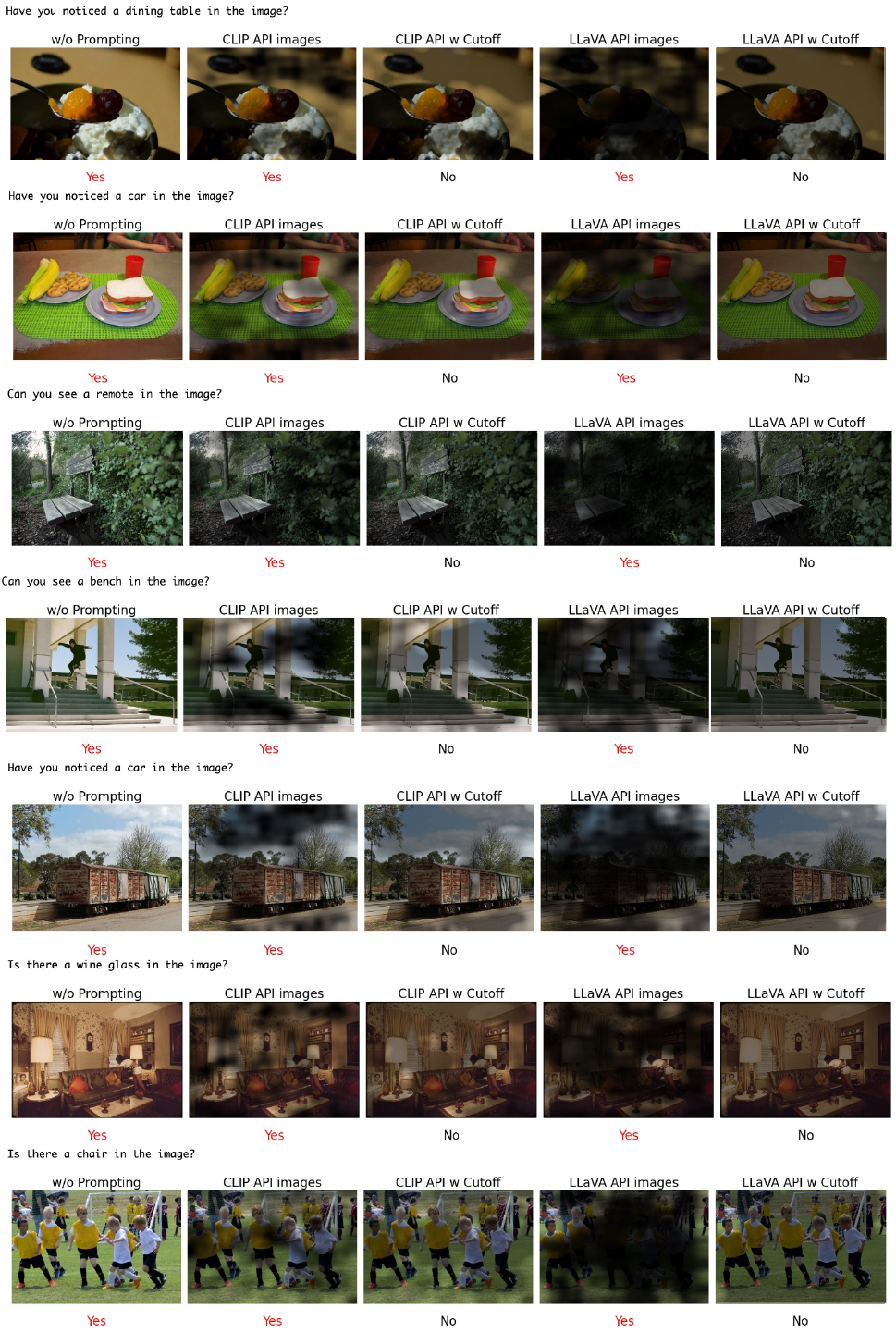}
    \caption{Examples of Successful Cases of Cutoff API Prompting of Objects Absent in Images.}
\end{figure}

\end{document}